\pdfoutput=1
\documentclass[11pt]{article}
\usepackage[preprint]{acl}
\usepackage{times}
\usepackage{latexsym}
\usepackage{microtype}
\usepackage{inconsolata}
\usepackage{amsmath}
\usepackage{amssymb}
\usepackage{booktabs}
\usepackage{array}
\usepackage{tabularx}
\usepackage{multirow}
\usepackage{graphicx}
\usepackage{enumitem}
\usepackage{xspace}
\usepackage{url}

\usepackage[textsize=tiny]{todonotes}
\def\BibTeX{{\rm B\kern-.05em{\sc i\kern-.025em b}\kern-.08em
    T\kern-.1667em\lower.7ex\hbox{E}\kern-.125emX}}

\title{SyRA-CS: Synthetic-to-Real Adaptation for Code-Switched ASR with Meta-Learning}
\author{Fadel Thior, Ahmer Tabassum, Haania Siddiqui, Hanan Aldarmaki, Thamar Solorio\\
Mohamed bin Zayed University of Artificially Intelligence \\
\texttt{fadel.thior@mbzuai.ac.ae}\\
}
\begin{document}
\maketitle

\begin{abstract}
Code-switched (CS) automatic speech recognition (ASR) remains difficult because real CS speech is scarce and synthetic speech does not fully match it. We investigate whether synthetic CS speech is more useful for learning an adaptation-friendly initialization than for ordinary continual fine-tuning. We propose SyRA-CS, a pair-specific synthetic-to-real adaptation framework based on Reptile meta-learning. For each language pair and pretrained model, we meta-train on the corresponding synthetic CS-FLEURS subset and then adapt using one hour of labeled real target speech. We evaluate four architectural settings: Encoder--CTC, FastConformer--RNN-T, Attention Encoder--Decoder (AED) and Speech-LLM. Experiments cover on Arabic-, French-, German-, and Hindi-English, against synthetic-only transfer, real-only fine-tuning, and continual fine-tuning. SyRA-CS improves over continual fine-tuning on WER in all 16 model--pair comparisons and on CER in 15 of 16, with relative macro WER reductions of 17.7\% for XLS-R and 5.3--5.8\% for the other three models. The gains hold across the tested models but vary in size across architectures, language pairs and metrics.

\end{abstract}

\section{Introduction}
Code-switching (CS)---the alternation of two or more languages within a conversation or utterance---is a common phenomenon in multilingual communities, but presents additional challenges for Automatic Speech Recognition (ASR) systems. The challenge is both linguistic and data-centric: real CS corpora are expensive to collect and transcribe, and switching patterns vary with speaker community, accent, discourse function, topic, and orthographic convention. As a result, ASR models trained for one CS condition often degrade under another \citep{sitaram2019survey,lyu2010seame,li2012mer}.

Synthetic CS speech offers an attractive way to cheaply expand supervision data \citep{yan2025csfleurs,dhawan2023unified}, but synthetic speech does not perfectly match real speech. Synthetic utterances may have cleaner acoustics, less speaker and accent variability. 
Therefore, simply fine-tuning on synthetic CS speech may overfit to synthetic regularities instead of generalizing to real CS recognition.
An alternative approach is meta-learning: a method that trains a model to \emph{become adaptable}, by repeatedly simulating small adaptation episodes and updating the parameters so that a few gradient steps on a new, data-scarce task lead to good performance \citep{finn2017maml}. The output is not a final recognizer but an initialization from which fine-tuning on a small labeled set is fast and stable.
This paper asks a focused question: \emph{under a limited real data budget, can a meta-learned initialization from synthetic CS speech adapt better to real CS speech compared to standard training strategies?} We use Reptile \citep{nichol2018firstorder} because it is a first-order meta-learning algorithm that repeatedly simulates adaptation episodes and moves the initialization toward parameters that are easy to reach after a few gradient steps:  
our contribution is not a new meta-learning optimizer or a benchmark of all meta-learning algorithms 
(Section~\ref{sec:why-reptile}). Instead, we test whether a lightweight meta-initializer is useful for synthetic-to-real CS ASR under severe target scarcity.
We train one Reptile initialization for each combination of language pair and pretrained model. Our experiments cover four architectural settings: XLS-R 300M with a CTC prediction
head \citep{babu2021xlsr}, Nemotron 3.5 with a
FastConformer--RNN-T architecture \citep{nemotron35modelcard},
Whisper-small with an attention encoder--decoder architecture \citep{whispersmallmodelcard}, and Nemotron-Labs-Audex-2B as a Speech-LLM \citep{audex2bmodelcard}. Within each model--pair combination, the training regimes start from
the same pretrained checkpoint and use the corresponding synthetic CS-FLEURS subset. Methods requiring real-domain adaptation use the same one-hour labeled target subset.
Hindi-English is evaluated on MUCS 2021
\citep{diwan2021mucs}, while Arabic-English, French-English,
and German-English are evaluated on SwitchLingua
\citep{xie2025switchlingua}. This design evaluates within-pair synthetic-to-real adaptation across different pretrained ASR systems; it does not learn a shared multilingual
meta-initialization.
Our contributions are:
\begin{itemize}[leftmargin=*,topsep=2pt,itemsep=2pt]

    \item We formulate pair-specific synthetic-to-real code-switched ASR as a data-efficient meta-learning problem, using synthetic speech to learn an initialization for adaptation with one hour of labeled real target speech 
    
    \item We investigate Reptile across four architectural settings: Encoder--CTC, FastConformer--RNN-T, attention encoder--decoder, and Speech-LLM, using comparisons
    against synthetic-only transfer, real-only fine-tuning, and continual fine-tuning within each pretrained model.
    
    \item We characterize the code-switching profile of each real target set (switching ratio) and assess the robustness of the gains to
  the two smallest evaluation sets.
\end{itemize}

\section{Related Work}
\subsection{Multilingual self-supervised}
Self-supervised speech learning has become the dominant paradigm for low-resource ASR. \citet{baevski2020wav2vec2} introduced wav2vec~2.0, which learns contextualized acoustic representations from raw unlabeled audio and can then be fine-tuned with limited transcribed speech. HuBERT improved masked prediction with cluster-based targets \citep{hsu2021hubert}, and XLS-R scaled the wav2vec~2.0 recipe to nearly half a million hours of multilingual audio across 128 languages \citep{babu2021xlsr}. Large multilingual speech resources such as FLEURS \citep{conneau2022fleurs}, Common Voice \citep{ardila2020commonvoice}, VoxPopuli \citep{wang2021voxpopuli}, and MMS \citep{pratap2023mms} have further strengthened multilingual transfer.

\subsection{Code-switched ASR and synthetic data}
A recent systematic review of end-to-end CS-ASR studies notes that, although data augmentation is widely used, it remains less explored than many model-centric approaches \citep{agro2025codeswitchingendtoendautomaticspeech}. In response, recent studies have used synthetic text and speech to improve low-resource code-switched ASR. Several works explore how synthetic code-switched data can be created or leveraged when real code-switched speech is scarce. \citet{nguyen2025can} show that ASR systems can be trained with synthetic phrase-level code-switching derived from monolingual resources, while \citet{nguyen2025asyncswitch} propose AsyncSwitch, which adapts models on large amounts of code-switched text before aligning this knowledge with limited paired speech. More broadly, \citet{hamed2025impact} find that the value of synthetic code-switched data depends strongly on its quality and the downstream task. At the same time, recent work has introduced stronger benchmarks and architectures for code-switched ASR. \citet{yan2025csfleurs} present CS-FLEURS, a multilingual benchmark covering 113 unique code-switched language pairs, and show that synthetic training data improves performance on both seen and unseen pairs. \citet{ugan2024decm} show that bilingual ASR systems still struggle on realistic code-switching benchmarks, while \citet{ye2024sc} propose a mixture-of-experts architecture to better handle language switching in streaming and non-streaming settings. Earlier work also explored semi-supervised and unified approaches \citep{tazakka2024indonesian,dhawan2023unified}. Overall, these works show that synthetic data is becoming an important tool for code-switched ASR, especially in low-resource settings where real code-switched speech is scarce. They also suggest that the main challenge is not only generating more data, but generating data that is linguistically and acoustically useful for adaptation.
\begin{figure*}[ht]
    \centering
    \includegraphics[width=\textwidth]{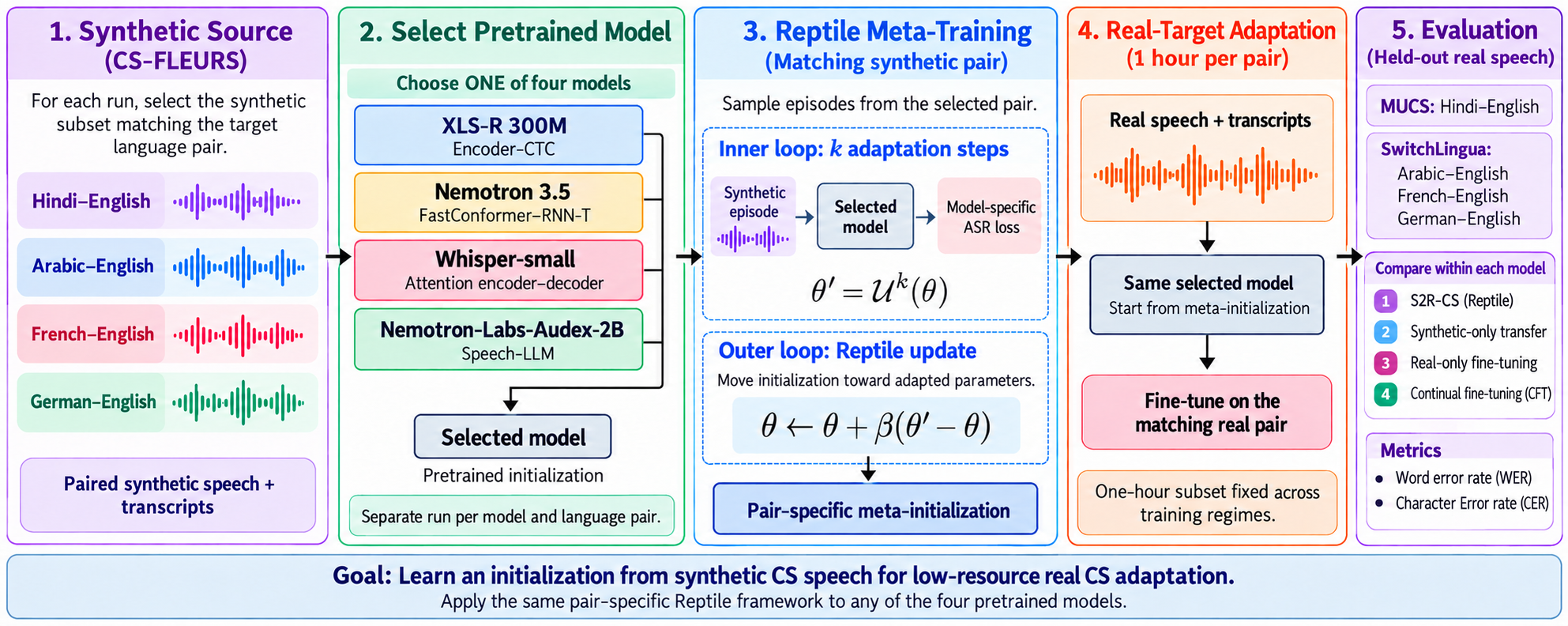}
    \caption{Overview of the Pair-Specific Synthetic-to-Real Reptile Adaptation Framework for Code-Switched ASR.}
    \label{fig:methodology}
\end{figure*}
\subsection{Meta-learning for ASR} 
Meta-learning is well suited to low-resource ASR because it explicitly learns parameters that can adapt quickly to new conditions with limited supervision. Early work applied MAML-style meta-learning to end-to-end ASR, learning
initializations from high-resource languages that adapt quickly to
low-resource targets \citep{hsu2020meta}, and \citet{winata2020meta}
extended this idea to code-switched speech with meta-transfer learning,
which draws episodes from monolingual corpora so that the initialization
transfers to limited real CS data. Our work differs in that episodes are
sampled from \emph{synthetic} CS speech of the same language pair and the
meta-learner is first-order Reptile rather than MAML. Recent ASR work has strengthened this direction: Meta-Adapter shows that meta-initialized adapters improve self-supervised speech adaptation for low-resource languages \citep{chen2024meta}, SMILE extends meta-learning to few-shot multilingual ASR through speech meta in-context learning \citep{hsu2024smile}, and MetaMixSpeech shows that task augmentation can improve generalization in multilingual low-resource ASR \citep{chen2025metamixspeech}. Together, these studies motivate meta-learning as an effective framework for fast speech adaptation under severe data scarcity. 
Related ideas also appear in multilingual NLP, where targeted adaptation has been shown to improve transfer to low-resource language families \citep{downey2024targeted}. In code-switched ASR, recent work such as AsyncSwitch emphasizes adaptation to limited paired code-switched speech after prior exposure to code-switched resources \citep{nguyen2025asyncswitch}.

\section{Data, Model, and Evaluation}
\subsection{Backbones and architecture scope}
\label{sec:backbones}

We evaluate SyRA-CS with four pretrained models spanning
different ASR architectures:
\begin{itemize}[leftmargin=*,topsep=2pt,itemsep=2pt]
    \item \textbf{XLS-R 300M (Encoder-CTC):}
    \texttt{wav2vec2-xls-r-300m}, with a linear CTC
    prediction head \citep{babu2021xlsr}.
    \item \textbf{Nemotron 3.5 (FastConformer--RNN-T):}
    a FastConformer encoder with an RNN-T decoder
    \citep{nemotron35modelcard}.  
    \item \textbf{Whisper-small (AED):}
    an attention-based Transformer encoder--decoder
    \citep{whispersmallmodelcard}. 
    \item \textbf{Nemotron-Labs-Audex-2B (Speech-LLM):}
    an audio-text language model used for speech
    transcription \citep{audex2bmodelcard}.
\end{itemize}

The synthetic CS-FLEURS data and real target data provide
supervised training and adaptation examples. They are not
used for additional self-supervised pretraining.

Let $a$ index the pretrained model and let $\theta$ denote
its trainable parameters. Reptile operates on these parameters
using the model-specific transcription objective:
\begin{equation}
\mathcal{L}_{a}=
\begin{cases}
\mathcal{L}_{\mathrm{CTC}},  & a=\text{XLS-R},\\
\mathcal{L}_{\mathrm{RNNT}}, & a=\text{Nemotron},\\
\mathcal{L}_{\mathrm{AR}},   & a\in\{\text{Whisper},\text{Audex}\},
\end{cases}
\end{equation}

CTC marginalizes over frame-level alignments compatible
with the transcript \citep{graves2006connectionist},
whereas RNN-T uses a transducer alignment objective.
For the autoregressive models, the transcription objective
is
\begin{equation}
\mathcal{L}_{\mathrm{AR}}(\theta;x,y)
=
-\sum_{u=1}^{|y|}
\log p_{\theta}(y_u\mid y_{<u},x,c_a),
\end{equation}
where $c_a$ denotes any model-specific task tokens or
transcription prompt. For the Speech-LLM, this expression
describes loss on the target transcription tokens. 
Across models, parameter counts, pretraining histories, tokenization, and decoding mechanisms
differ. The cross-model experiments therefore examine whether the benefit of Reptile extends across architectural settings; absolute differences between models cannot be attributed to architecture alone.

\subsection{Source tasks: CS-FLEURS data}
Our synthetic source data are drawn from CS-FLEURS \citep{yan2025csfleurs}, a corpus of synthetic code-switched speech spanning multiple language pairs. We use four code-switched subsets from XTTS version as it contains training data for our meta learning experiments: Hindi-English (6 hours), French-English (10.28h), Arabic-English (4.51 hours)
 and German-English (9.37 hours), where each subset corresponds to one language pair and the reported durations denote the total amount of synthetic training audio available for that pair. For a model $a$ and target pair $p$, synthetic episodes are sampled only from the corresponding source subset $S_p$. Thus, each meta-training run learns a model-specific,
pair-specific initialization rather than an initialization
trained jointly across language pairs.

\subsection{Target tasks: MUCS and Switchlingua data}
We evaluate adaptation under real-world code-switched conditions using all four pairs as target tasks. Real-world Hindi-English speech data is from MUCS \citep{diwan2021mucs}, French-English (2.64 hours), Arabic-English (1.43 hours) and German-English (1.40 hours) are from the SwitchLingua dataset \citep{xie2025switchlingua}. SwitchLingua provides real audio recordings from bilingual
speakers, but the spoken prompts are generated code-switched text samples. Thus, the target audio is human speech, while the linguistic content follows a synthetic-script collection protocol. Unlike SwitchLingua,  MUCS Hindi-English is extracted from human spoken tutorials. The speech is therefore real
instructional audio, and the code-switching primarily arises from technical lecture content rather than from generated code-switched text. For each target pair, we randomly select a 1-hour labeled subset for adaptation and evaluate final performance on the corresponding real code-switched test set. As CS speech data is scarce, this setup is intentionally low-resource and is designed to measure how effectively a model trained on synthetic code-switched speech can transfer to real speech with only a small amount of target-domain supervision.
\subsection{Metrics and Text normalization}
We report Word Error Rate (WER) and Character Error Rate (CER) using Levenshtein edit distance \citep{levenshtein1966binary}:
\begin{equation}
\begin{aligned}
\mathrm{WER}&=\frac{S+D+I}{N_w},\\
\mathrm{CER}&=\frac{S_c+D_c+I_c}{N_c}.
\end{aligned}
\end{equation}
The same normalization is applied to references and hypotheses for all systems before scoring: unicode normalization, removal of markup artifacts, punctuation stripping, whitespace normalization, and case normalization for Latin-script text. We do not introduce system-specific post-processing. For Arabic-English, the evaluation uses the same common normalization pipeline and does not apply additional Arabic-specific orthographic collapsing unless explicitly configured in the released scoring script. WER and CER are standard and comparable across all four pairs; we complement them with a description of the code-switching profile of each test set. 
Multi-reference WER \citep{7404847} and PolyWER \citep{kadaoui-etal-2024-polywer} are not used here because MUCS and SwitchLingua provide a single reference per utterance; constructing alternative references manually would introduce subjective normalization choices.

\section{Meta Learning Adaptation}
Let
$\mathcal{P}=\{\text{ara-eng},\text{fra-eng},
\text{deu-eng},\text{hin-eng}\}$
and
$\mathcal{A}=\{\mathrm{XLSR},\mathrm{Nemotron},
\mathrm{Whisper},\mathrm{Audex}\}$.
For each pair $p\in\mathcal{P}$, let $S_p$ denote its
synthetic CS-FLEURS source set, $A_p$ its one-hour real
adaptation set, and $E_p$ its held-out real evaluation set.
For each model $a\in\mathcal{A}$, we learn a separate
initialization $\theta_{a,p}^{\mathrm{meta}}$ starting from
that model's pretrained initialization $\theta_{a,0}$.
The symbol $\theta$ denotes the parameters updated during
training. 
At meta-iteration $t$, each synthetic episode $i$ starts
from
$\theta_{a,p,i}^{(0)}=\theta_{a,p,t}$.
It performs $k_a$ inner updates using the model-specific
loss from Section~\ref{sec:backbones}:
\begin{equation}
\begin{aligned}
\theta_{a,p,i}^{(j+1)}
&=\theta_{a,p,i}^{(j)}
-\alpha_a
\nabla_{\theta}
\mathcal{L}_{a}
\bigl(\theta_{a,p,i}^{(j)};B_{p,i}^{(j)}\bigr),
\end{aligned}
\end{equation}
where $B_{p,i}^{(j)}\subset S_p$ is a synthetic minibatch
and $j=0,\ldots,k_a-1$. This notation expresses a gradient
step; the optimizer used in each implementation is reported
with the experimental configuration.

Let
$\widetilde{\theta}_{a,p,i}
=\theta_{a,p,i}^{(k_a)}$.
For a meta-batch of $m_a$ episodes, the Reptile update is
\begin{equation}
\begin{aligned}
\theta_{a,p,t+1}
&=\theta_{a,p,t}\\
&\quad+
\frac{\beta_{a,t}}{m_a}
\sum_{i=1}^{m_a}
\bigl(\widetilde{\theta}_{a,p,i}-\theta_{a,p,t}\bigr),
\end{aligned}
\label{eq:reptile}
\end{equation}
where $\beta_{a,t}$ is the outer step size. The update uses
parameter displacements without differentiating through
the inner optimization trajectory.

After meta-training, the initialization is adapted to the
one-hour real subset:
\begin{equation}
\begin{aligned}
\phi_{a,p}^{(0)}
&=\theta_{a,p}^{\mathrm{meta}},\\
\phi_{a,p}^{(u+1)}
&=\phi_{a,p}^{(u)}
-\eta_a\nabla_{\phi}
\mathcal{L}_{a}
\bigl(\phi_{a,p}^{(u)};B_p^{(u)}\bigr),
\end{aligned}
\end{equation}
where $B_p^{(u)}\subset A_p$.
The resulting model $\phi_{a,p}^{\star}$ is evaluated on
$E_p$. Neither meta-training nor target adaptation uses
evaluation examples.

\subsection{Why First-Order Meta-Learning?}
\label{sec:why-reptile}
Reptile is particularly suitable for our setting because it avoids differentiating through the inner optimization trajectory
\citep{nichol2018firstorder}. In MAML, after $k$ inner updates
$\theta'_{\tau}=U_{\tau}^{k}(\theta)$, the meta-update requires
\[
\theta \leftarrow \theta -
\beta\nabla_{\theta}
\mathcal{L}^{\mathrm{out}}_{\tau}
\left(U_{\tau}^{k}(\theta)\right),
\]
and therefore backpropagates through the sequence of inner-loop optimization steps. In its standard form, this introduces second-order derivatives and substantial memory and computational overhead. This overhead is particularly relevant when repeatedly adapting large pretrained speech models, including CTC, transducer, encoder--decoder, and Speech-LLM systems. Reptile instead uses the parameter displacement
\[
\theta \leftarrow
\theta+\beta(\theta'_{\tau}-\theta),
\]
without differentiating through the trajectory that produced $\theta'_{\tau}$. The outer update can therefore be applied across our four architectural settings while the inner loop uses each model's transcription objective. Our hypothesis is that repeated adaptation on synthetic CS episodes produces an initialization that transfers more effectively to limited real target speech than ordinary synthetic-domain fine-tuning. This hypothesis is tested through within-model comparisons against CFT. Reptile's first-order formulation reduces the derivative overhead associated with standard second-order MAML, but does not remove the cost of repeated inner-loop training.

\section{Experiments and Results}
\subsection{Baselines and Reproducibility}
For each model $a$ and language pair $p$, we compare four
training regimes:
\begin{itemize}[leftmargin=*,topsep=2pt,itemsep=2pt]
    \item \textbf{Synthetic-only transfer:}
    train from the model's pretrained checkpoint on $S_p$
    and evaluate directly on $E_p$, without real adaptation.
    
    \item \textbf{Real-only fine-tuning:}
    fine-tunes the pretrained model on $A_p$ only.
    
    \item \textbf{Continual fine-tuning (CFT):}
    train first on $S_p$ and then adapt on $A_p$.
    
    \item \textbf{SyRA-CS:}
    learn a Reptile initialization from episodes sampled
    from $S_p$, then adapt on $A_p$.
\end{itemize}
Within each model--pair combination, all regimes start
from the same pretrained checkpoint. Methods using real
adaptation share the same one-hour subset, and all methods
are evaluated on the same held-out set. Trainable parameter scope, tokenization, transcript normalization, and inference settings are fixed across regimes within each model. CFT is the closest baseline because it uses the same
synthetic source set and real target subset as SyRA-CS.
Their comparison tests the value of episodic Reptile
training relative to ordinary sequential training.
Table~\ref{tab:hyperparams} reports the Reptile configuration
for each model. 

\begin{table}[t]
\centering
\small
\setlength{\tabcolsep}{3pt}
\resizebox{\columnwidth}{!}{%
\begin{tabular}{lll}
\toprule
\textbf{Setting}
& \textbf{XLS-R 300M}
& \shortstack[l]{\textbf{Nemotron 3.5 /}\\
                 \textbf{Whisper-small / Audex-2B}} \\
\midrule
Architecture
& Encoder--CTC
& \shortstack[l]{FastConformer--RNN-T /\\
                 AED / Speech-LLM} \\
Trainable parameters & Full model & Full model \\
Meta-iterations & 500 & 500 \\
Inner steps $k_a$ & 50 & 50 \\
Inner learning rate $\alpha_a$
& $3\times10^{-4}$ & $2\times10^{-5}$ \\
Outer step size $\beta_a$ & 0.1 & 0.1 \\
Batch size & 2 & 2 \\
Target fine-tuning epochs & 50 & 50 \\
GPU & $1\times$ RTX 4090 & RTX A4500 \\
\midrule
\multicolumn{3}{l}{Setting: model-specific, pair-specific Reptile} \\
\multicolumn{3}{l}{Pairs: ara-eng, fra-eng, deu-eng, hin-eng} \\
\multicolumn{3}{l}{Source: corresponding CS-FLEURS subset} \\
\multicolumn{3}{l}{Target: 1h per pair; MUCS and SwitchLingua} \\
\bottomrule
\end{tabular}%
}
\caption{Reptile configurations. The grouped column
reports shared settings; architectures follow the
model order.}
\label{tab:hyperparams}
\end{table}

\subsection{Main results}
\label{sec:main-results}
\begin{figure*}[t]
\centering
\includegraphics[width=\textwidth]
{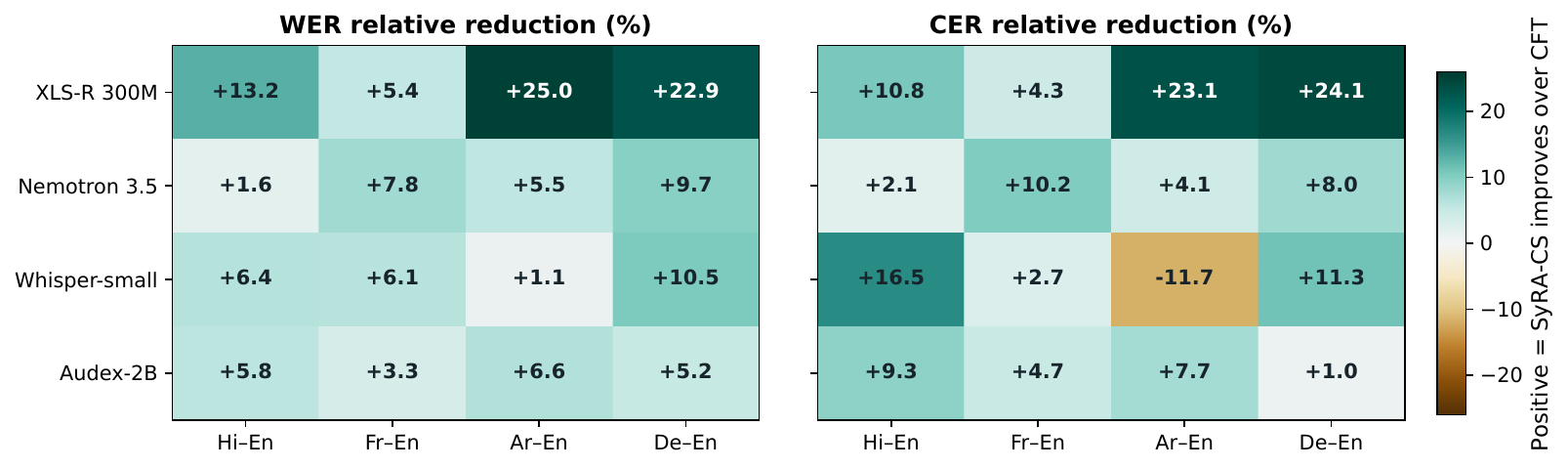}
\caption{Relative WER and CER reductions of SyRA-CS
against CFT, computed as
$100(e_{\mathrm{CFT}}-e_{\mathrm{SyRA}})/
e_{\mathrm{CFT}}$.
Positive values indicate improvement; negative values
indicate higher error.
}
\label{fig:relative-all}
\end{figure*}

Table~\ref{tab:main_results} compares the four training regimes within each pretrained model, and 
SyRA-CS improves over CFT in all 16 model--pair WER comparisons and in 15 of the 16 CER comparisons, and attains the lowest error among the four regimes in 31 of the 32 model--pair--metric cells. 
Macro averages over the four pairs are computed from the rounded scores in Table~\ref{tab:main_results}; relative macro reductions are computed from the macro error rates rather than by averaging the pair-level reductions in Figure~\ref{fig:relative-all}.

\begin{table*}[t]
\centering
\small
\setlength{\tabcolsep}{10pt}
\resizebox{\textwidth}{!}{%
\begin{tabular}{llcccccccc}
\toprule
\multirow{2}{*}{\textbf{Training regime}}
& \multirow{2}{*}{\textbf{Real data}}
& \multicolumn{2}{c}{\textbf{Hindi-English}}
& \multicolumn{2}{c}{\textbf{French-English}}
& \multicolumn{2}{c}{\textbf{Arabic-English}}
& \multicolumn{2}{c}{\textbf{German-English}} \\
\cmidrule(lr){3-4}
\cmidrule(lr){5-6}
\cmidrule(lr){7-8}
\cmidrule(lr){9-10}
& & \textbf{WER $\downarrow$} & \textbf{CER $\downarrow$}
  & \textbf{WER $\downarrow$} & \textbf{CER $\downarrow$}
  & \textbf{WER $\downarrow$} & \textbf{CER $\downarrow$}
  & \textbf{WER $\downarrow$} & \textbf{CER $\downarrow$} \\
\midrule

\multicolumn{10}{l}{\textbf{XLS-R 300M (Encoder--CTC)}} \\
Synthetic-only transfer & 0h
& 0.770 & 0.590 & 0.814 & 0.364
& 0.976 & 0.527 & 0.830 & 0.333 \\
Real-only fine-tuning & 1h
& 0.980 & 0.960 & 0.531 & 0.182
& 0.827 & 0.367 & 0.784 & 0.219 \\
CFT from synthetic & 1h
& 0.479 & 0.288 & 0.370 & 0.117
& 0.615 & 0.212 & 0.432 & 0.137 \\
SyRA-CS (ours) & 1h
& \textbf{0.416} & \textbf{0.257}
& \textbf{0.350} & \textbf{0.112}
& \textbf{0.461} & \textbf{0.163}
& \textbf{0.333} & \textbf{0.104} \\
\midrule

\multicolumn{10}{l}{
\textbf{Nemotron 3.5 (FastConformer--RNN-T)}} \\
Synthetic-only transfer & 0h
& 0.554 & 0.490 & 0.277 & 0.160
& 0.538 & 0.319 & 0.337 & 0.145 \\
Real-only fine-tuning & 1h
& 0.531 & 0.472 & 0.233 & 0.130
& 0.504 & 0.320 & 0.309 & 0.147 \\
CFT from synthetic & 1h
& 0.488 & 0.426 & 0.219 & 0.118
& 0.454 & 0.269 & 0.267 & 0.113 \\
SyRA-CS (ours) & 1h
& \textbf{0.480} & \textbf{0.417}
& \textbf{0.202} & \textbf{0.106}
& \textbf{0.429} & \textbf{0.258}
& \textbf{0.241} & \textbf{0.104} \\
\midrule

\multicolumn{10}{l}{\textbf{Whisper-small (AED)}} \\
Synthetic-only transfer & 0h
& 0.831 & 0.675 & 0.278 & 0.133
& 0.613 & 0.321 & 0.305 & 0.124 \\
Real-only fine-tuning & 1h
& 0.614 & 0.494 & 0.172 & 0.082
& 0.374 & 0.188 & 0.189 & 0.080 \\
CFT from synthetic & 1h
& 0.533 & 0.418
& 0.163 & 0.073
& 0.352 & \textbf{0.162}
& 0.191 & 0.080 \\
SyRA-CS (ours) & 1h
& \textbf{0.499} & \textbf{0.349}
& \textbf{0.153} & \textbf{0.071}
& \textbf{0.348} & 0.181
& \textbf{0.171} & \textbf{0.071} \\
\midrule

\multicolumn{10}{l}{
\textbf{Nemotron-Labs-Audex-2B (Speech-LLM)}} \\
Synthetic-only transfer & 0h
& 0.766 & 0.640 & 0.194 & 0.110
& 0.721 & 0.496 & 0.324 & 0.147 \\
Real-only fine-tuning & 1h
& 0.525 & 0.460
& 0.136 & 0.074
& 0.607 & 0.427 & 0.199 & 0.107 \\
CFT from synthetic & 1h
& 0.520 & 0.431 
& 0.121 & 0.064
& 0.577 & 0.443
& 0.193 & 0.102 \\
SyRA-CS (ours) & 1h
& \textbf{0.490} & \textbf{0.391} 
&\textbf{0.117} & \textbf{0.061}
& \textbf{0.539} & \textbf{0.409}
& \textbf{0.183} & \textbf{0.101} \\
\bottomrule
\end{tabular}%
}
\caption{Synthetic-to-real CS ASR performance across four pretrained models. WER and CER are reported as fractions. Real data denotes labeled target speech
used for adaptation.}
\label{tab:main_results}
\end{table*}

\paragraph{Largest gains for XLS-R.}
The encoder--CTC model benefits most: macro WER falls from 0.4740 to 0.3900 (17.7\%) and macro CER from 0.1885 to 0.1590 (15.6\%). The largest pair-level gains are on Arabic-English (WER 0.615 $\rightarrow$ 0.461) and German-English (0.432 $\rightarrow$ 0.333). XLS-R also has the weakest CFT baseline of the four models, leaving the most room for improvement.
 
\paragraph{Smaller but consistent gains for the other models.}
Nemotron 3.5 and Audex-2B improve both metrics on every pair, with macro WER/CER reductions of 5.3\%/4.4\% and 5.8\%/7.5\%, respectively. Audex gains most on Arabic-English (WER 0.577 $\rightarrow$ 0.539) and Hindi-English (0.520 $\rightarrow$ 0.490). Whisper-small improves WER on every pair (5.5\% macro) and macro CER by 8.3\%, driven mainly by Hindi-English (CER 0.418 $\rightarrow$ 0.349). Its Arabic-English result is the only regression: WER decreases slightly (0.352 $\rightarrow$ 0.348) while CER increases (0.162 $\rightarrow$ 0.181), so pair-level conclusions can depend on the metric.

\paragraph{Variation across models and language pairs.}
Figure~\ref{fig:relative-all} shows pair-level relative reductions against CFT. WER reductions range from 5.4\% to 25.0\% for XLS-R, 1.6\% to 9.7\% for Nemotron, 1.1\% to 10.5\% for Whisper, and 3.3\% to 6.6\% for Audex. All four models improve both WER and CER on Hindi-English, French-English, and German-English. On Hindi-English, WER improves by 13.2\%, 1.6\%, 6.4\%, and 5.8\% for XLS-R, Nemotron, Whisper, and Audex, respectively. Because Hindi-English differs from the SwitchLingua pairs in corpus, domain, and switching density (Section~\ref{sec:cs-analysis}) simultaneously, these results cannot isolate switching density as a driver of the gains.
\paragraph{Other baselines.}
Synthetic-only transfer yields the highest error in 28 of 32 cells, confirming a substantial synthetic-to-real gap. CFT outperforms real-only fine-tuning in 29 of 32 cells (exceptions: Whisper German-English WER, a tie on Whisper German-English CER, and Audex Arabic-English CER), making it a strong baseline. Real-only XLS-R nearly fails on Hindi-English (WER 0.980), which inflates its macro score, so we treat CFT as the primary comparison. Because CFT and SyRA-CS use the same synthetic subset and the same one-hour real subset, their gap reflects the training strategy rather than additional data.
\subsection{Code-switching profile and Robustness checks}
\label{sec:cs-analysis}

\begin{figure}[t]
\centering
\includegraphics[width=\columnwidth]
{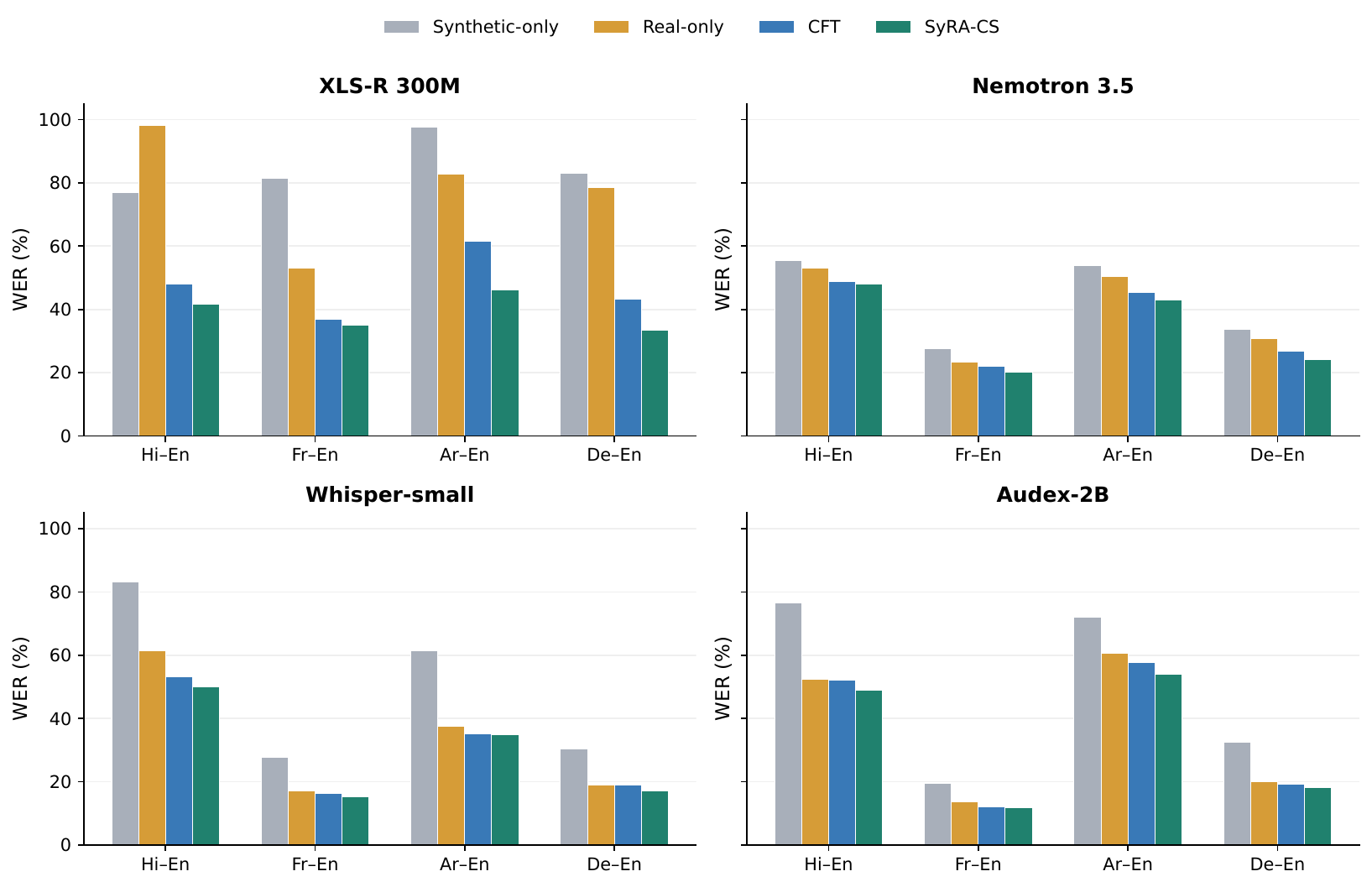}
\caption{Pair-level WER for all four models and training
regimes. Values are percentages; lower is better.}
\label{fig:pair-wer}
\end{figure}

\begin{figure}[t]
\centering
\includegraphics[width=\columnwidth]
{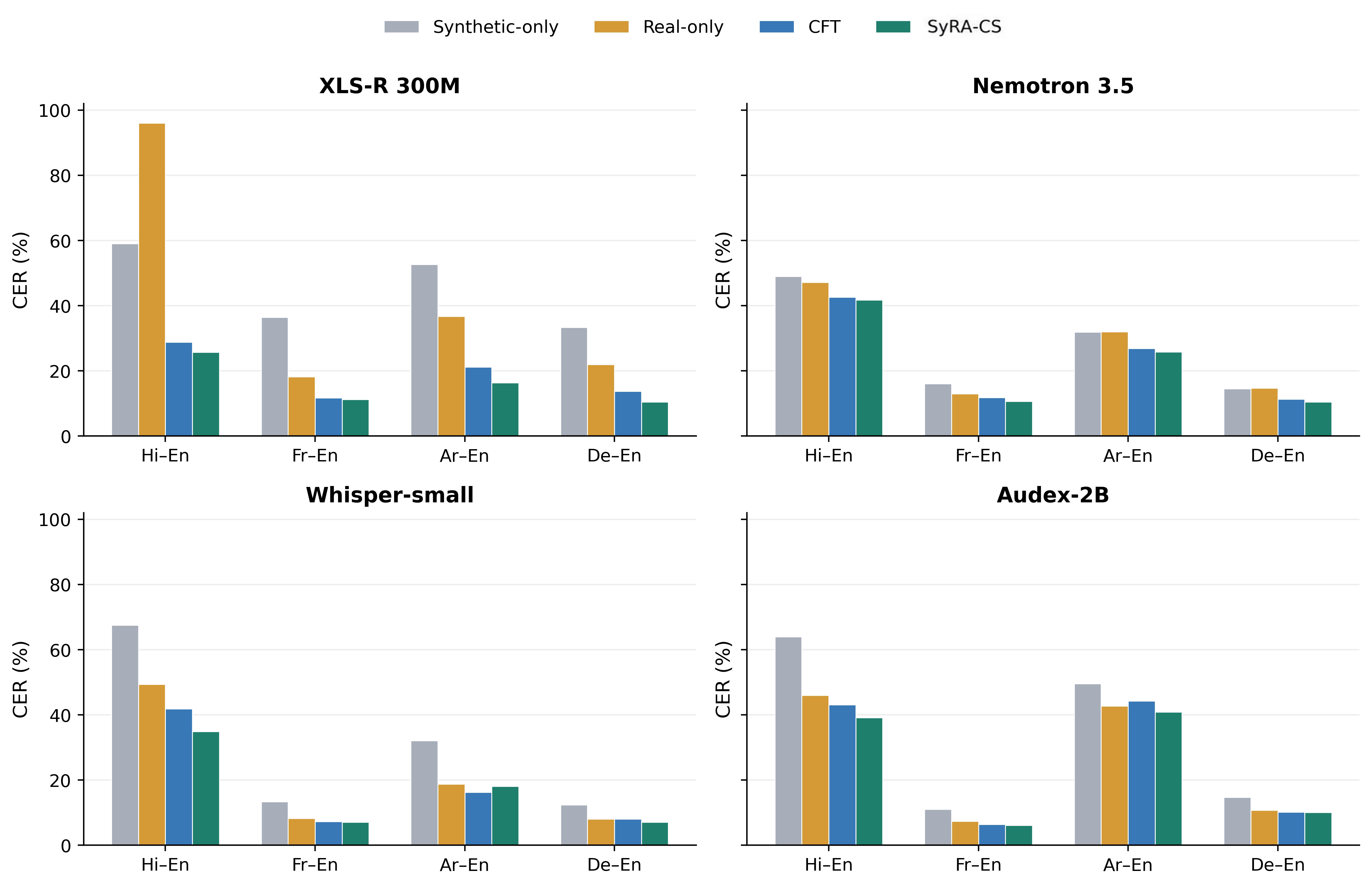}
\caption{Pair-level CER for all four models and training
regimes. Values are percentages; lower is better.}
\label{fig:pair-cer}
\end{figure}
\begin{figure}[t]
\centering
\includegraphics[width=\columnwidth]{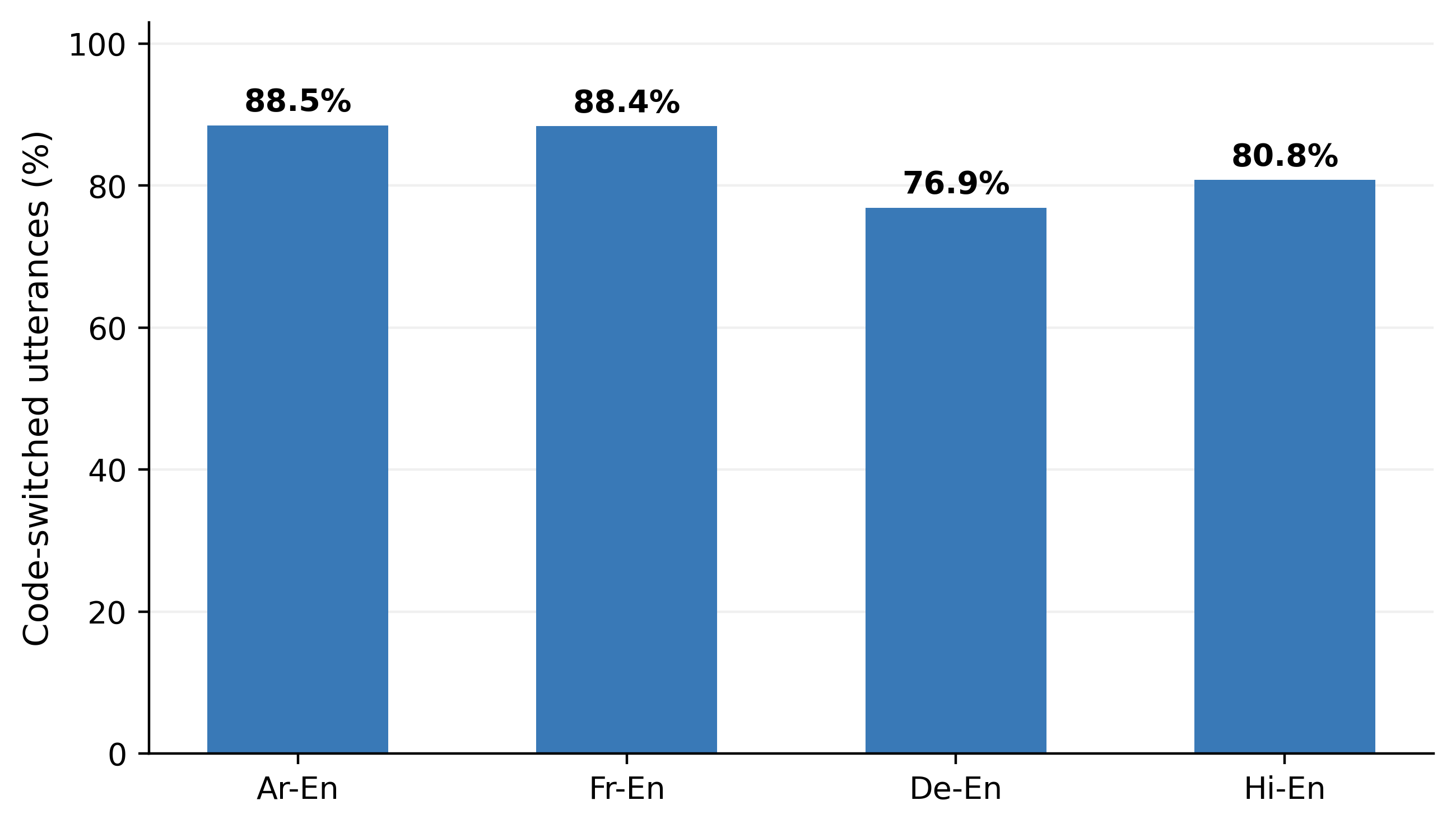}
\caption{Code-switching ratio in each target test set.}
\label{fig:cs-ratio-by-pair}
\end{figure}

Figure~\ref{fig:cs-ratio-by-pair} shows that code-switched utterances account for 88.5\% of Arabic-English, 88.4\% of French-English, 76.9\% of German-English, and 80.8\% of Hindi-English test utterances. Hindi-English has the highest reported switching density, at 2.996 switches per utterance.
These statistics establish substantial mixed-language coverage, but do not identify which linguistic phenomena account for a model's improvements.

\begin{figure}[t]
\centering
\includegraphics[width=\columnwidth]
{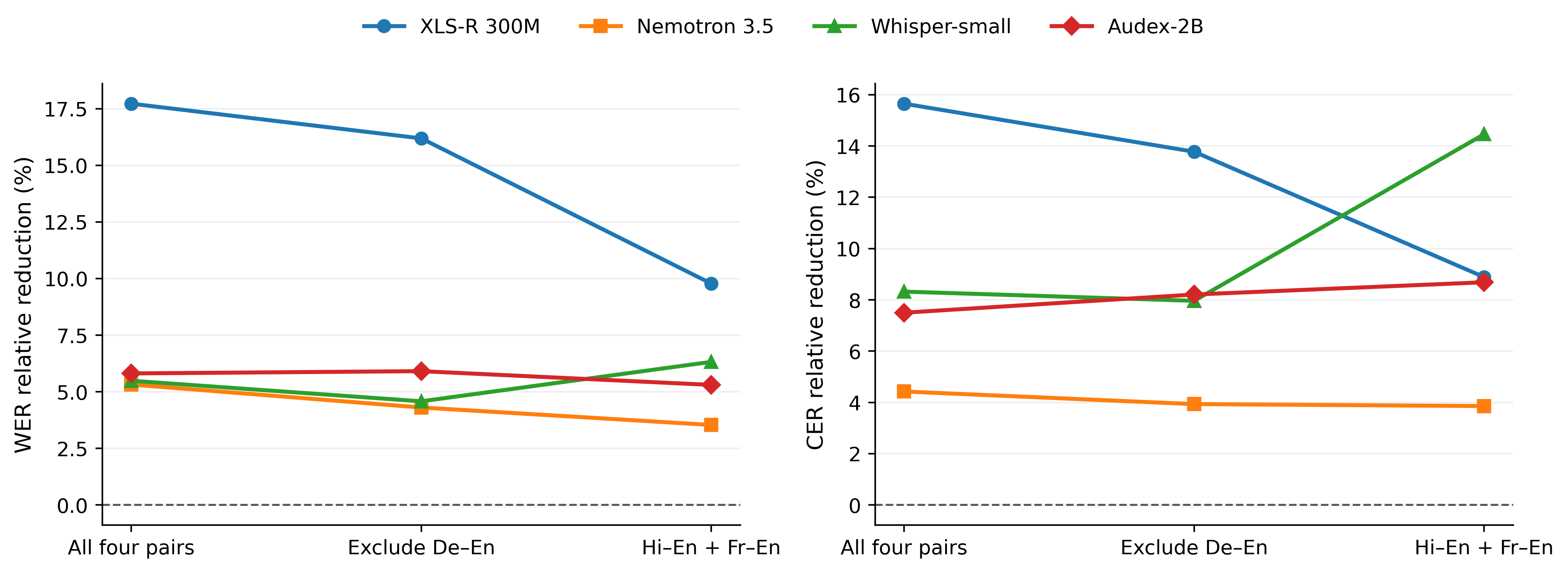}
\caption{Sensitivity of relative macro WER/CER reduction
to the retained language pairs. Positive values favor SyRA-CS. Lines connect subset summaries; they do not represent a time trend or uncertainty intervals.Each retained pair receives equal weight.}
\label{fig:subset-sensitivity}
\end{figure}

\paragraph{Robustness to small evaluation sets.}
\label{sec:small-test-sensitivity}
German-English and Arabic-English contain only 1.40 and 1.43 hours of real speech before the one-hour adaptation split, leaving 104 and 218 test utterances. To assess dependence on these sets, we recompute the CFT--SyRA-CS macro comparison after excluding German-English and then after excluding both pairs (Figure~\ref{fig:subset-sensitivity}). 
Without German-English, relative macro WER reductions are 16.2\% for XLS-R, 4.3\% for Nemotron, 4.6\% for Whisper, and 5.9\% for Audex, with CER reductions of 13.8\%, 3.9\%, 8.0\%, and 8.2\%. Retaining only Hindi-English and French-English, WER reductions are 9.8\%, 3.5\%, 6.3\%, and 5.3\%, and CER reductions are 8.9\%, 3.9\%, 14.5\%, and 8.7\%. All four models therefore keep positive macro gains on both metrics in every subset. XLS-R's gain shrinks as its two strongest pairs are removed, whereas Whisper's CER gain grows once its Arabic-English regression is excluded.
These exclusions are descriptive sensitivity checks, not significance tests. All averages weight pairs equally; pooled error rates and paired uncertainty estimates require utterance-level edit statistics, and variability across training seeds and adaptation subsets is not assessed here.

\section{Discussion}
The results support the central hypothesis: synthetic CS speech is more effective when used to learn an adaptation-friendly initialization than when used only through ordinary sequential fine-tuning. This is shown by the consistent SyRA-CS improvements over CFT, the closest matched baseline. The largest Reptile benefit appears where the pretrained model is furthest from a usable CS recognizer. XLS-R is pretrained only with self-supervision and its CTC head is learned from scratch, so the initialization carries the most weight: real-only fine-tuning nearly fails on Hindi-English (WER 0.980), and SyRA-CS yields the largest gain (17.7\% macro WER). Nemotron, Whisper, and Audex already have supervised ASR pretraining, start from stronger CFT baselines, and gain less (5.3--5.8\%). This suggests that a meta-learned initialization matters most when the backbone lacks task-specific supervision, although a single model per architecture prevents a causal claim. Whisper's Arabic-English result does not contradict this picture. Its WER improves while its CER worsens, a pattern that indicates character-level edits \emph{within} otherwise correctly recognized
words rather than additional word errors; this is the signature of Arabic orthographic variation that our shared, script-agnostic normalization does not
collapse. We therefore read the regression as a normalization artifact rather than a failure of the initialization. 
Practically, SyRA-CS uses no data beyond what CFT uses, making it a low-cost alternative under a one-hour real-data budget, provided the additional meta-training compute is acceptable.
\section{Conclusion}
We introduced SyRA-CS, a pair-specific Reptile framework for synthetic-to-real CS ASR adaptation. Across four pretrained models spanning encoder--CTC, FastConformer--RNN-T, AED, and Speech-LLM architectures, and with the same synthetic source data and one-hour real adaptation budget within each model, SyRA-CS lowers WER relative to continual fine-tuning in all 16 model--pair comparisons and CER in 15, with relative macro WER reductions of 5.3\% to 17.7\%. Gains are largest for XLS-R, smaller but consistent for the other three models, and remain positive when the two smallest evaluation sets are excluded. These results are consistent with the intended role of Reptile: learning a lightweight initialization that specializes readily to real CS speech. Future work will add compute-matched controls, multi-seed significance testing, CS diagnostics for all models, additional language pairs and  more spontaneous speech corpora. 
\section*{Limitations}
 This study has four main limitations. 
\paragraph{Pair-specific adaptation.}Reptile is applied in a
pair-specific setting rather than as a shared multilingual meta-learner. 
Thus, the experiments evaluate within-pair synthetic-to-real adaptation
$p\rightarrow p$, not a shared initialization
$\theta_{a,p}^{\mathrm{meta}}$ trained jointly over all
$\{S_p\}_{p\in\mathcal{P}}$ 
for cross-pair
meta-transfer. The results therefore support pair-specific Reptile adaptation, not universal multilingual meta-initialization.

\paragraph{Unexplored synthetic and real data requirements.} We evaluate SyRA-CS using fixed synthetic-source datasets and a one-hour labeled real adaptation budget, without systematically varying either data volume. The experiments therefore do not establish the minimum amount of synthetic or real speech needed to achieve a given level of recognition accuracy, nor how these two data sources trade off during adaptation.
Controlled data-scaling experiments are needed to
characterize these requirements and identify diminishing returns from additional data.

\paragraph{Model coverage and attribution of gains.}
Although the study includes four architectural
settings, each is represented by a single pretrained
model. Differences in model capacity, pretraining,
tokenization, trainable parameter scope, and optimization prevent attributing differences in
Reptile's effectiveness to architecture alone. Gains are also not uniform (Whisper's Arabic-English CER increases under SyRA-CS). Finally, we do not compare Reptile with alternative meta-learning algorithms such as MAML or its first-order variant, so we cannot determine whether the benefit is specific to Reptile.

\paragraph{Statistical reliability.}
All results come from a single training seed and a single random one-hour
adaptation subset per pair. German-English and Arabic-English retain only
104 and 218 test utterances after the adaptation split, so several
pair-level differences (e.g., Nemotron Hindi-English WER, $+1.6\%$;
Whisper Arabic-English WER, $+1.1\%$) are likely within seed and subset
variance. The subset analysis in Section~\ref{sec:small-test-sensitivity} is a
descriptive sensitivity check, not a significance test; paired
utterance-level statistics and multi-seed runs are needed before the
smaller gains can be considered reliable.


\bibliography{custom}
\end{document}